\documentclass[11pt]{article}
\usepackage{coling2018}
\usepackage{times}
\usepackage{url}
\usepackage{latexsym}
\usepackage{wrapfig}
\usepackage{graphicx}
\usepackage{subcaption}
\usepackage{amsmath}
\usepackage{ctable}
\usepackage{multirow}
\DeclareMathOperator*{\argmax}{arg\,max}

\title{A Dataset for Building Code-Mixed Goal Oriented Conversation Systems}

 \author{Suman Banerjee,  \quad Nikita Moghe, \quad Siddhartha Arora \quad \and \enspace Mitesh M. Khapra \\
        Indian Institute of Technology Madras, India \\ \{\normalsize \tt suman, nikitavam, sidarora, miteshk\}@cse.iitm.ac.in}
\date{}
\begin{document}
\maketitle
\begin{abstract}
There is an increasing demand for goal-oriented conversation systems which can assist users in various day-to-day activities such as booking tickets, restaurant reservations, shopping, \textit{etc}. Most of the existing datasets for building such conversation systems focus on monolingual conversations and there is hardly any work on multilingual and/or code-mixed conversations. Such datasets and systems thus do not cater to the multilingual regions of the world, such as India, where it is very common for people to speak more than one language and seamlessly switch between them resulting in code-mixed conversations. For example, a Hindi speaking user looking to book a restaurant would typically ask, ``Kya tum is \textit{restaurant} mein ek \textit{table book} karne mein meri \textit{help} karoge?" (``Can you help me in booking a table at this restaurant?"). 
To facilitate the development of such code-mixed conversation models, we build a goal-oriented dialog dataset containing code-mixed conversations. 
Specifically, we take the text from the DSTC2 restaurant reservation dataset and create code-mixed versions of it in Hindi-English, Bengali-English, Gujarati-English and Tamil-English. We also establish initial baselines on this dataset using existing state of the art models. This dataset along with our baseline implementations is made publicly available for research purposes. 


\end{abstract}
\blfootnote{
    %
    %
    %
    %
    %
    %
    \hspace{-0.65cm}  
    This work is licensed under a Creative Commons 
    Attribution 4.0 International License.
    License details:
    \url{http://creativecommons.org/licenses/by/4.0/}
}
\section{Introduction}
\label{intro}
Over the past few years, there has been an increasing demand for virtual assistants which can help users in a wide variety of tasks in several domains such as entertainment, finance, healthcare, e-commerce, \textit{etc.} 
To cater to this demand, several commercial conversation systems such as Siri, Cortana, Allo have been developed. While these systems are still far from general purpose open domain chat, they perform reasonably well for certain goal-oriented tasks such as setting alarms/reminders, booking appointments, checking movie show timings, finding directions for navigation, \textit{etc}. Apart from these commercial systems, there has also been significant academic research to advance the state of the art in conversation systems \cite{NRM,NCM,attn_with_int,movieLi,personaNCM,VHRED}. Most of this academic research is driven by publicly available datasets such as Twitter conversation dataset \cite{ritter2010}, Ubuntu dialog dataset \cite{ubuntu}, Movie subtitles dataset \cite{movie_sub} and DSTC2 restaurant reservation dataset \cite{dstc2}. In this work, we focus on goal-oriented conversations such as the ones contained in the DSTC2 dataset.

Most of the datasets and state of the art systems mentioned above are monolingual. Specifically, all the utterances and responses in the conversations are in one language (typically, English) and there are no multilingual and/or code-mixed utterances/responses. However, in several multilingual regions of the world, such as India, it is natural for speakers to produce utterances and responses which are multilingual and code-mixed. For example, Table \ref{example} shows real examples of how bilingual speakers from India talk when requesting someone to help them reserve a restaurant or book movie tickets. As it can be seen, when engaging in such informal conversations it is very natural for such speakers to use code-mixed utterances, mixing their native language with English. Apart from India, such code-mixing is also prevalent in other multilingual regions of the world, for example, Spanglish (Spanish-English), Frenglish (French-English), Porglish (Portuguese-English) and so on. 
To cater to such users, it is essential to create datasets containing code-mixed conversations and thus facilitate the development of code-mixed conversation systems. 

With the above motivation, we build a dataset containing code-mixed goal-oriented conversations for four Indian languages. Specifically, we take every utterance from the DSTC2 restaurant reservation dataset and ask a mix of in-house and crowdsourced workers to create a corresponding code-mixed utterance involving their native language and English. We simply instructed the workers to (i) assume that they were chatting with a friend who spoke the same native language as them in addition to English, (ii) not try very hard to translate the sentence completely to their native language but feel free to switch to English whenever they wanted (just as they would in a normal conversation with a friend) and (iii) use Romanized text instead of the native language's script. The resulting dataset contains utterances of the type shown in Table \ref{example}. We found that 87.73\% of the created utterances were code-mixed, 7.18\% had only English words and 5.09\% had only native language words. The four languages that we chose were Hindi, Bengali, Tamil and Gujarati which have 422M, 83M, 60M and 46M native speakers respectively. 

Apart from reporting various statistics about this data (such as CM-index \cite{CMI} and I-index \cite{I-index}), we also report some initial baselines by evaluating some state of the art approaches on the proposed dataset. Specifically, we evaluate a standard sequence-to-sequence model with an attention mechanism \cite{seq2seq} and a hierarchical recurrent encoder-decoder model \cite{HRED}. Our code implementing these models along with the dataset is available freely for research purposes\footnote{ \footnotesize \url{https://github.com/sumanbanerjee1/Code-Mixed-Dialog}}. To the best of our knowledge, this is the first conversation dataset containing code-mixed conversations and will hopefully enable further research in this area. In particular, since the data is 5-way parallel (English, Bengali, Hindi, Tamil, Gujarati) it would be useful for building jointly trained code-mixed models.  

\begin{table}[]
\centering
\small{
\begin{tabular}{ll}
\hline
\multicolumn{1}{c|}{Languages} & \multicolumn{1}{c}{Utterances}                                                                                                                                                                                      \\\hline
\multicolumn{1}{c|}{English}                       & \begin{tabular}[c]{@{}l@{}}
\textbf{Speaker 1}: Hi, Can you help me in booking a table at this restaurant?\\ 
\textbf{Speaker 2}: Sure, would you like something in cheap, moderate or expensive price range?
\end{tabular}              \\\hline
\multicolumn{1}{c|}{Hindi-English}                 & \begin{tabular}[c]{@{}l@{}}
\textbf{Speaker 1}: \textit{Hi}, kya tum is \textit{restaurant} mein ek \textit{table book} karne mein meri \textit{help} karoge?\\ 
\textbf{Speaker 2}: \textit{Sure}, kya aap \textit{cheap}, \textit{moderate} ya \textit{expensive price range} mein kuch \textit{like} karenge?
\end{tabular} \\\hline
\multicolumn{1}{c|}{Bengali-English}               & \begin{tabular}[c]{@{}l@{}}
\textbf{Speaker 1}: \textit{Hi}, tumi ki ei \textit{restaurant} ey ekta \textit{table book} korte \textit{help} korbe amake?\\ 
\textbf{Speaker 2}: \textit{Sure}, aapni ki \textit{cheap, moderate} na \textit{expensive price range} ey kichu \textit{like} korben ?
\end{tabular}     \\\hline
\multicolumn{1}{c|}{English}                       & \begin{tabular}[c]{@{}l@{}}
\textbf{Speaker 1}: Hello, can you tell me about the show timings of ``Black Panther"?\\ 
\textbf{Speaker 2}: Sure, would you like to book tickets for today or any other day?
\end{tabular}              \\\hline
\multicolumn{1}{c|}{Gujarati-English}              & \begin{tabular}[c]{@{}l@{}}
\textbf{Speaker 1}: \textit{Hello}, mane Black Panther na \textit{show timings} janavo.\\ 
\textbf{Speaker 2}: \textit{Sure}, shu tame aaj ni ke koi anya divas ni \textit{ticket book} karva mango cho?
\end{tabular}                                                                                                                                                     \\\hline
\multicolumn{1}{c|}{Tamil-English}                 &           \begin{tabular}[c]{@{}l@{}}
\textbf{Speaker 1}: \textit{Hello}, ``Black Panther" \textit{show timings} eppo epponu solla mudiuma\\ 
\textbf{Speaker 2}: Kandipa, \textit{tickets} innaiku \textit{book} pannanuma illana vera ennaikku?
\end{tabular}      \\\hline          
\end{tabular}
\caption{\label{example}Example code-mixed utterances in the specified languages.}
}
\end{table}

%
%
    %
    %
    %
    %
    %
    %

\section{Related Work}
\newcite{survey_on_dialog_datasets} report an excellent (and up-to-date) survey of existing dialog datasets. For brevity, we only mention some of the important points from their survey and refer the reader to the original paper for more details. To begin with, we note that existing dialog datasets can be categorized along 3 main dimensions. The first dimension is the modality of the dataset, \textit{i.e.}, whether the dataset contains spoken conversations \cite{switchboard,dstc5} or text conversations \cite{NPS,ritter2010,ubuntu}. The second dimension is whether the dataset contains goal-oriented conversations or open-ended conversations. A goal-oriented conversation typically involves chatting for the sake of completing a task such as the Dialog State Tracking Challenge (DSTC) datasets which involve tasks for reserving a restaurant \cite{dstc2}, checking bus schedules \cite{dstc1}, collecting tourist information \cite{dstc3} and so on. Such datasets are also typically domain-specific. Open-ended conversations on the other hand involve general chat on any topic and there is no specific end task. Some popular examples of datasets containing such open-ended conversations are the Ritel Corpus \cite{ritel}, NPS Chat Corpus \cite{NPS}, Twitter Corpus \cite{ritter2010}, \textit{etc}. The third dimension is whether the dataset contains human-human conversations or human-bot conversations. As the name suggests, human-bot conversation datasets contain conversations between humans and an existing conversation system (typically a domain-specific goal-oriented bot) \cite{dstc1,dstc2,dstc3}. 
Human-human conversations, on the other hand, can contain spontaneous conversations between humans, as are typically observed in discussion forums \cite{argument}, chat rooms \cite{NPS}, SMS messages \cite{NUS} and so on. Human-human conversations can also contain scripted dialogs such as scripts of movies \cite{movieDIC}, TV shows \cite{tvd}, \textit{etc}. It is surprising that of the 63 conversation datasets developed in the past \cite{survey_on_dialog_datasets}, none contain multilingual conversations. In particular, none of them contain code-mixed conversations from multilingual regions of the world. There is clearly a need to fill this gap and we believe that the dataset developed as a part of this work is a small step in that direction.

In general, the research community has been interested in developing datasets, tools and approaches for code-mixed content. This interest is largely triggered by the abundance of code-mixed content found in chats, emails, social media platforms, \textit{etc}. In the context of such code-mixed content, existing works have looked at the problems of language identification \cite{langID4,langID1,langID3,langID2},  part-of-speech tagging \cite{POS1,POS2,POS3}, user profiling \cite{user_profilling}, topic modeling \cite{topic_modelling}, information retrieval \cite{IR} and language modeling \cite{lang_model,lang_model2,lang_model3}.  However, to the best of our knowledge, ours is the first work on developing code-mixed conversation systems for goal-oriented dialogs.

\begin{table}[]
\centering
\small{
\begin{tabular}{cl}
\hline
\multicolumn{1}{c|}{User}                                          &\multicolumn{1}{c}{Bot}                                                                                                                                                                            \\ \hline
 \multicolumn{1}{l|}{1. \textless SILENCE\textgreater}                  & \begin{tabular}[c]{@{}l@{}}Hello, welcome to the Cambridge restaurant system.\\ You can ask for restaurants by area, price range or food type. \\ How may I help you?\end{tabular} \\\hline
 \multicolumn{1}{l|}{2. i'm looking for a moderately priced restaurant.} & What kind of food would you like?                                                                                                                                                    \\\hline
 \multicolumn{1}{l|}{3. it doesn't matter.}                              & What part of town do you have in mind?                                                                                                                                               \\\hline
 \multicolumn{1}{l|}{4. the south part of town.}                        & api\_call R\_cuisine south moderate                                                                                                                                                   \\\hline

\multicolumn{2}{l}{\qquad pizza\_hut\_cherry\_hinton R\_post\_code pizza\_hut\_cherry\_hinton\_post\_code}                                                                                                                                                       \\
\multicolumn{2}{l}{\qquad pizza\_hut\_cherry\_hinton R\_cuisine italian}                                                                                                                                                                                         \\
\multicolumn{2}{l}{\qquad pizza\_hut\_cherry\_hinton R\_location south}                                                                                                                                                                                          \\
\multicolumn{2}{l}{\qquad pizza\_hut\_cherry\_hinton R\_phone pizza\_hut\_cherry\_hinton\_phone}                                                                                                                                                                 \\
\multicolumn{2}{l}{\qquad pizza\_hut\_cherry\_hinton R\_address pizza\_hut\_cherry\_hinton\_address}                                                                                                                                                             \\
\multicolumn{2}{l}{\qquad pizza\_hut\_cherry\_hinton R\_price moderate}                                                                                                                                                                                          \\
\multicolumn{2}{l}{\qquad pizza\_hut\_cherry\_hinton R\_rating 3}                                                                                                                                                                                                \\
\multicolumn{2}{l}{\qquad restaurant\_alimentum R\_post\_code restaurant\_alimentum\_post\_code}                                                                                                                                                                 \\
\multicolumn{2}{l}{\qquad restaurant\_alimentum R\_cuisine european}                                                                                                                                                                                             \\
\multicolumn{2}{l}{\qquad restaurant\_alimentum R\_location south}                                                                                                                                                                                               \\
\multicolumn{2}{l}{\qquad restaurant\_alimentum R\_phone restaurant\_alimentum\_phone}                                                                                                                                                                           \\
\multicolumn{2}{l}{\qquad restaurant\_alimentum R\_address restaurant\_alimentum\_address}                                                                                                                                                                       \\
\multicolumn{2}{l}{\qquad restaurant\_alimentum R\_price moderate}                                                                                                                                                                                               \\
\multicolumn{2}{l}{\qquad restaurant\_alimentum R\_rating 10}                                                                                                                                                                                                    \\\hline
 \multicolumn{1}{l|}{5. \textless SILENCE\textgreater}                  & \begin{tabular}[c]{@{}l@{}}restaurant\_alimentum is a nice restaurant \\ in the south of town serving modern\_european food.\end{tabular}                                              \\\hline
 \multicolumn{1}{l|}{6. can i get the address?}                        & \begin{tabular}[c]{@{}l@{}}Sure, restaurant\_alimentum is on \\ restaurant\_alimentum\_address.\end{tabular}                                                                          \\\hline
 \multicolumn{1}{l|}{7. and whats the phone number for that?}           & \begin{tabular}[c]{@{}l@{}}The phone number of restaurant\_alimentum \\ is restaurant\_alimentum\_phone.\end{tabular}                                                                  \\\hline
 \multicolumn{1}{l|}{8. okay, thank you, good bye.}                       & you are welcome.                                                                     \\\hline                                                                                                 
\end{tabular}
}
\caption{\label{example_chat_en} An example chat from the English version of DSTC2 dataset  \cite{weston2017}.}
\end{table}

\section{Background: DSTC2 Restaurant Reservation Dataset}
We build on top of the goal-oriented restaurant reservation dialog dataset which was released as part of the second Dialog State Tracking Challenge (DSTC2) \cite{dstc2}. This dataset contains conversations between crowdsourced workers and existing dialog systems (bots). Specifically, the workers were asked to book a table at a restaurant with the help of a bot. These dialog systems consisted of modules like automatic speech recognizer, natural language interpreter, dialog manager, response generator and a speech synthesizer \cite{young2000}. The dialog manager used polices which were either hand-crafted or learned by formulating the problem as a partially observable Markov decision process (POMDP) \cite{POMDP}. The speech input from the user was first converted to text and then fed to the dialog system. For this, the authors used two Automatic Speech Recognition (ASR) modules out of which one was artificially degraded in order to simulate noisy environments. 
The workers could request for restaurants based on 3 slots: \textit{area} (5 possible values), \textit{cuisine} (91 possible values) and \textit{price range} (3 possible values). The workers were also instructed to change their goals and look for alternative \textit{areas},\textit{ cuisines} and \textit{price ranges} in the middle of the dialog. This was done to account for the unpredictability in natural conversations. 
The conversations were then transcribed and the utterances were labeled with different dialog states. For example, each utterance was labeled with its semantic intent representation (\textit{request[area], inform[area = north]}) and the dialog turns were labeled with annotations such as constraints on the slots (\textit{cuisine = italian}), requested slots (\textit{requested = \{phone, address\}}) and the method of search (\textit{by\_constraints, by\_alternatives}). Such annotations are useful for domain-specific slot-filling based dialog systems.

\newcite{weston2017} argued that for various domains collecting such explicit annotations for every state in the dialog is tedious and expensive. Instead, they emphasized on building end-to-end dialog systems (as opposed to slot-filling based systems) by adapting this dataset and treating it as a simple sequence of utterance-response pairs (without any explicit dialog states associated with the utterances). In addition, the authors also created API calls which can be issued to an underlying Knowledge Base (KB) and appended the resultant KB triples to each dialog. Table \ref{example_chat_en} shows one small sample dialog from this adapted dataset along with the API calls. Notice that the API call uses the information of all the constraints specified by the user so far and then receives all triples from the restaurant KB which match the user's requirements. This dataset facilitated the development of models \cite{weston2017,QRN,HCN,manning} which just predict the bot utterances and API calls without explicitly tracking the slots. 
Table \ref{enstats} reports the statistics of this dataset. In this work, we create code-mixed versions of this dataset in 4 different languages as described below.

\begin{table}
\centering
\small{
\begin{tabular}{ll}
\hline
\# of Utterances                            & 49167   \\
\# of Unique utterances 				& 6733 \\
Average \# of utterances per dialog       & 15.19  \\
Average \# of words per utterance       & 7.71 \\
Average \# of words per dialog            & 120.33 \\
Average \# of KB triples per dialog &38.24\\
\# of Train Dialogs                     & 1168 \\
\# of Validation Dialogs                     & 500 \\
\# of Test Dialogs                     & 1117 \\
Vocabulary size                                &1229     \\
\hline 
\end{tabular}
\caption{\label{enstats}
Statistics of the English version of DSTC2 dataset}}
\end{table}

\section{Code-Mixed Dialog Dataset}
In this section, we describe the process used for creating a new dataset containing code-mixed conversations. Specifically, we describe (i) the process used for extracting unique utterance templates from the original DSTC2 dataset, (ii) the process of creating code-mixed translations of these utterances with the help of in-house and crowdsourced workers and (iii) the process used for evaluating the collected conversations. Finally, we report some statistics about the dataset.

\subsection{Extracting Unique Utterance Templates}
\label{utt_template}
We found that many utterances in the original English version of DSTC2 dataset (henceforth referred to as En-DSTC2) have the same sentence structure but only differ in the values of the \textit{areas}, \textit{cuisines}, \textit{price ranges} and entities such as \textit{restaurant names, addresses, phone numbers} and \textit{post codes}. For example, consider these two sentences which only differ in the \textit{area} and \textit{cuisine}: (i) ``Sorry, there is no \textit{chinese} restaurant in the \textit{north} part of town." and (ii) ``Sorry, there is no \textit{italian} restaurant in the \textit{west} part of town". Both these sentences can be thought of as instantiations of the generic template: ``Sorry, there is no [CUISINE] restaurant in the [AREA] part of town." wherein the placeholders [AREA] and [CUISINE] get replaced by different values. We used the KB provided by \newcite{weston2017} to find all the entities appearing in all the utterances and replaced them by placeholders such as: [AREA], [CUISINE], [PRICE], [RESTAURANT], [ADDRESS], [PHONE] and [POST\_CODE]. Further, since the authors had mentioned that the KB provided was not perfect/complete, we did some manual inspection to find all such entities and came up with a list of 536 such entity words. After replacing these words with their respective placeholders we obtained 3590 unique English utterances.

\subsection{Creating Code-Mixed Translations}
According to \newcite{myers} code-mixing involves a native language which provides the morphosyntactic frame and a foreign language whose \textit{linguistic units} such as \textit{phrases, words} and \textit{morphemes} are inserted into this morphosyntactic frame. The native language is called the \textit{Matrix} while the foreign language is called the \textit{Embedding}. Our work focuses on creating a conversation dataset wherein 4 different Indian languages, \textit{viz.}, Hindi, Bengali, Gujarati and Tamil serve as the \textit{Matrix} and English serves as the \textit{Embedding}. We used a mix of in-house and crowdsourced workers to create a code-mixed version of the original DSTC2 dataset. For example, for Hindi and Gujarati, we did not have enough in-house speakers so we completely relied on crowdsourcing for creating the data but then used in-house workers to verify the collected data. For Bengali, all the data was created by in-house annotators who were native Bengali speakers and proficient in English. Lastly, for Tamil, roughly 40\% of the data was created with the help of crowdsourced workers and the rest with the help of in-house workers. Irrespective of whether the workers were crowdsourced or in-house we used the same set of instructions as described below.

We instructed the annotators to assume that they were chatting with a friend who is a native speaker of Hindi (or Gujarati, Bengali, Tamil) but also speaks English well (typically, because English was the language in which the friend did most of his/her education). To explain the idea of code-mixing, we showed them example utterances where it was natural for the user to mix English words while chatting in the native language. They were then shown an English utterance from the DSTC2 dataset and asked to create its code-mixed translation in the native language keeping the above code-mixed examples in mind. They were asked to use Roman script irrespective of whether the word being used belongs to English or the native language (in particular, they were clearly instructed to not use the native language's script). As expected, we observed that while translating, the annotators tend to retain some difficult-to-translate and colloquially relevant English words as it is. The annotators were also clearly instructed to refrain from producing pure translations (\textit{i.e.}, they were asked to not try hard to translate English words which they would typically not translate in an informal conversation). Also, the annotators were instructed to retain the placeholder words ([AREA], [CUISINE], \textit{etc.}) as it is and not translate them.

We used Amazon Mechanical Turk (AMT) as the platform for crowdsourcing. Each Human Intelligence Task (HIT) required the user to give code-mixed translations of 5 utterances and was priced at \$0.2.  Once we collected the code-mixed translations of all the utterance templates that were extracted using the procedure described in the previous subsection, we then instantiated them into proper sentences by replacing the placeholders ([AREA], [CUISINE], \textit{etc.}) with the corresponding entities as present in the original DSTC2 dataset. For every dialog in the original DSTC2 dataset, every utterance was then replaced by its code-mixed translation resulting in an end-to-end code-mixed conversation.

\begin{table}[]
\begin{minipage}{.55\linewidth}
\centering
\small {\begin{tabular}{@{}ll|ll@{}}
\hline
\textbf{}                                                                                              & \textbf{}              & \textbf{\begin{tabular}[c]{@{}l@{}}In-house\\ Workers\end{tabular}} & \textbf{Evaluators} \\ \hline
\textbf{Avg. Age}                                                                                      & \textbf{}              & 25.2                                                                & 24.6                \\ \hline
\multirow{2}{*}{\textbf{Gender}}                                                                       & \textbf{Female}        & 33.3\%                                                              & 25.0\%              \\  
                                                                                                       & \textbf{Male}          & 66.7\%                                                              & 75.0\%              \\ \hline
\multirow{3}{*}{\textbf{Highest Education}}                                                            & \textbf{Undergraduate} & 25.0\%                                                              & 33.3\%              \\ 
                                                                                                       & \textbf{Graduate}      & 41.7\%                                                              & 33.3\%              \\ 
                                                                                                       & \textbf{Postgraduate}  & 33.3\%                                                              & 33.3\%              \\ \hline
\multirow{2}{*}{\textbf{\begin{tabular}[c]{@{}l@{}}English Medium \\ Schooling\end{tabular}}}          & \textbf{Yes}           & 100\%                                                               & 100\%               \\ 
                                                                                                       & \textbf{No}            & 0\%                                                                 & 0\%                 \\ \hline
\multirow{3}{*}{\textbf{\begin{tabular}[c]{@{}l@{}}Frequency of \\ English usage\end{tabular}}}        & \textbf{Frequently}    & 75.0\%                                                              & 91.7\%              \\ 
                                                                                                       & \textbf{Occasionally}  & 25.0\%                                                              & 8.3\%               \\ 
                                                                                                       & \textbf{Rarely}        & 0\%                                                                 & 0\%                 \\ \hline
\multirow{3}{*}{\textbf{\begin{tabular}[c]{@{}l@{}}Frequency of\\ native language \\usage\end{tabular}}} & \textbf{Frequently}    & 100\%                                                               & 91.7\%              \\ 
                                                                                                       & \textbf{Occasionally}  & 0\%                                                                 & 8.3\%               \\ 
                                                                                                       & \textbf{Rarely}        & 0\%                                                                 & 0\%                 \\ \hline
\end{tabular}
\caption{\label{demograph} Demographic details of the in-house workers and the human evaluators.}
}
\end{minipage}\hfill
\begin{minipage}{.42\linewidth}
\centering
\small{
\begin{tabular}{@{}cccc@{}}
\hline
Datasets        & \rotatebox[origin=c]{45}{Colloquialism} & \rotatebox[origin=c]{45}{Intelligibility} & \rotatebox[origin=c]{45}{Coherent} \\ \hline
Hi-DSTC2             &4.20               &4.06              &4.21          \\
Be-DSTC2           &4.07               &4.05              &4.11          \\
Gu-DSTC2          &3.66               &3.60              &3.76          \\
Ta-DSTC2             &4.17               &3.96              &3.93         \\\hline
\end{tabular}
\caption{\label{evalstats} Average human ratings for different metrics.}
}
\end{minipage}
\end{table}

\subsection{Evaluating the Collected Dataset}
We did evaluations at two levels. The first evaluation was at the level of utterances wherein if the code-mixed translation of an utterance was obtained via crowdsourcing, then we got this translation verified by in-house evaluators. The evaluators were asked to check if (i) the translation was faithful to the source sentence, (ii) the code-mixing was natural and not forced and (iii) all translations used Roman script and not the native language's script. Any utterance which was flagged as erroneous by the evaluator was again crowdsourced and a new translation was solicited from AMT workers. If a worker's utterances were flagged erroneous often then we barred him/her from doing any more tasks. 

As mentioned in the previous section, once we collected such verified translations for all the utterance templates, we instantiated them and created complete end-to-end  dialogs containing  code-mixed utterances. Once the entire dialog was constructed, we conducted a separate human evaluation wherein we asked 12 in-house evaluators (3 evaluators per language) to read 100 code-mixed dialogs (entire dialogs as opposed to just some utterances) from each language and rate them on three metrics namely colloquialism, intelligibility and coherence on a scale of 1 (very poor) to 5 (very good) as defined below.

\begin{itemize}
\item \textbf{Colloquialism:} To check if the code-mixing was colloquial throughout the dialog and not forced.
\item \textbf{Intelligibility:} To check if the entire dialog could be easily understood by a bilingual speaker who could speak the native language as well as English.
\item \textbf{Coherence:} To check if the entire dialog looked coherent even though it was constructed by stitching together utterances which were independently translated and code-mixed (\textit{i.e.}, while translating an utterance annotators did not know what their preceding and following utterances were).
\end{itemize}
These 100 dialogs were chosen randomly from across the entire dataset for each language. The evaluators used for this were different from the in-house annotators used to create the original translations in order to reduce the bias in evaluations. The average ratings given by the evaluators for each of the languages are shown in Table \ref{evalstats} and are encouraging. The demographic details of the in-house workers and evaluators are shown in Table \ref{demograph}.

\subsection{Dataset Statistics and Analysis}
For every word in the code-mixed corpus, we were able to identify whether it was a word from the native language or English or language agnostic (named entities). It was easy to do this because we had the vocabulary of the original English DSTC2 corpus as well as named entities (so any word which was not in the original DSTC2 vocabulary or a named entity was a word from the native language). We also manually verified this list of words marked as native words and corrected discrepancies if any (\textit{i.e.}, we ensured that all the words which were marked as native words were actually native words). Note that the cuisine names such as \textit{Australian, Italian, etc.} have their own dedicated words in the native language. Table \ref{cmstats} summarizes various statistics about the dataset such as total vocabulary size, native language vocabulary size, \textit{etc}. We refer to the original English dataset as En-DSTC2 and the Hindi, Bengali, Tamil and Gujarati code-mixed datasets created as a part of this work as Hi-DSTC2, Be-DSTC2, Ta-DSTC2 and Gu-DSTC2 respectively. Below, we make a few observations from Table \ref{cmstats}.

\begin{figure}[!tbp]
        \centering
        \begin{subfigure}[b]{0.4\textwidth}
            \centering
            \includegraphics[width=0.92\textwidth ]{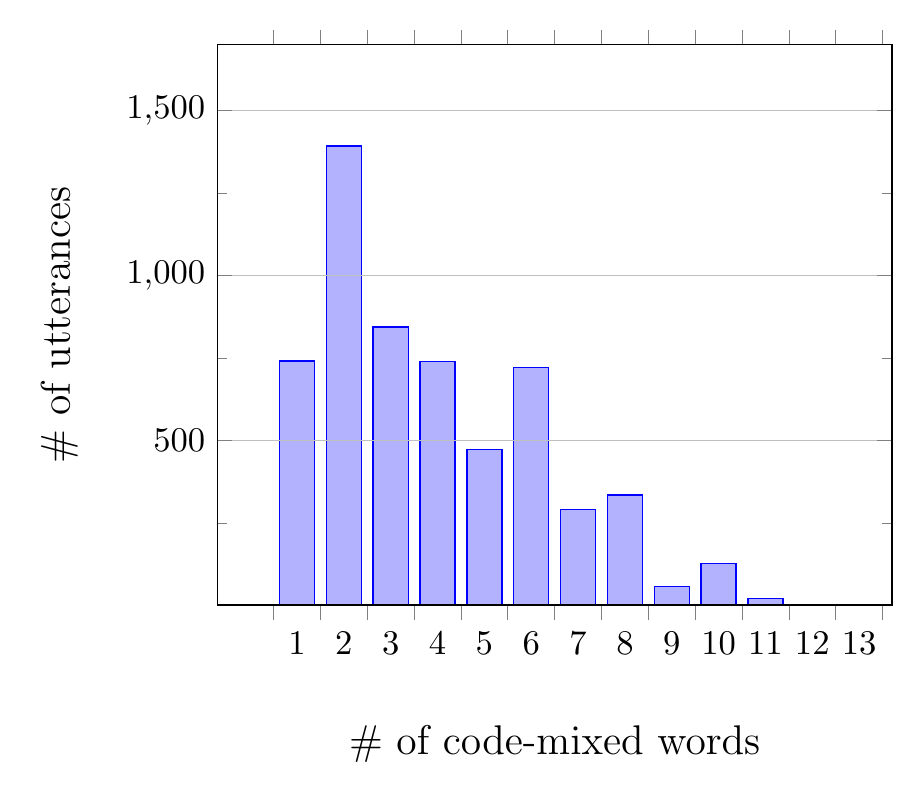}
            \caption[]%
            {{\small Hindi}}    
            \label{hin_hist}
        \end{subfigure}
        \begin{subfigure}[b]{0.4\textwidth}  
            \centering 
            \includegraphics[width=0.92\textwidth]{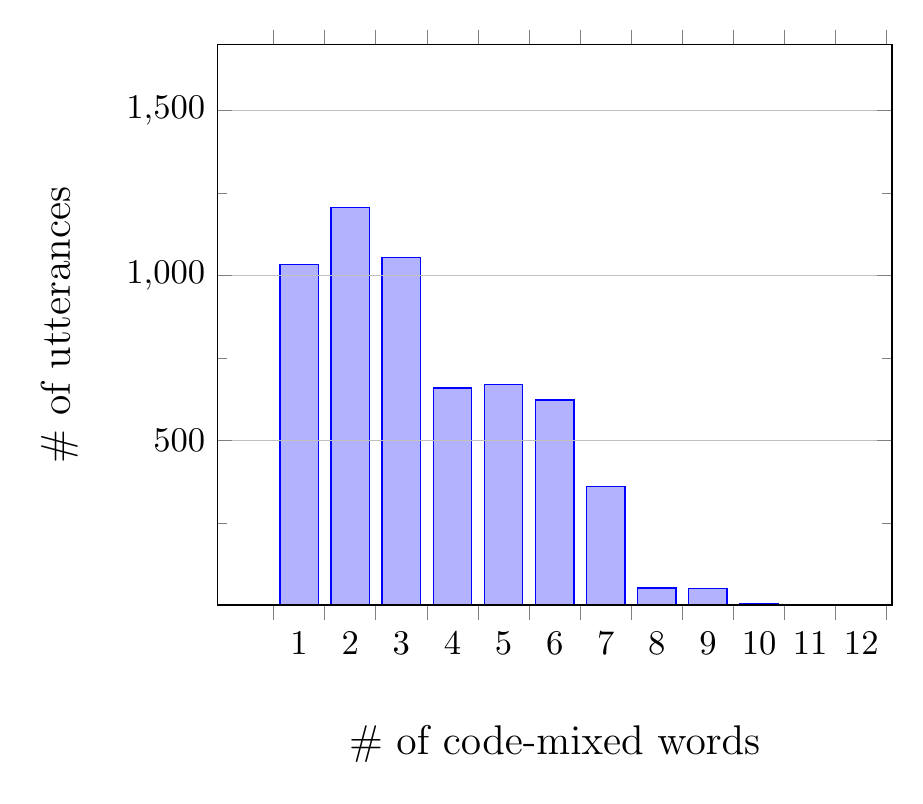}
            \caption[]%
            {{\small Bengali}}    
            \label{ben_hist}
        \end{subfigure}
        \begin{subfigure}[b]{0.4\textwidth}   
            \centering 
            \includegraphics[width=0.92\textwidth]{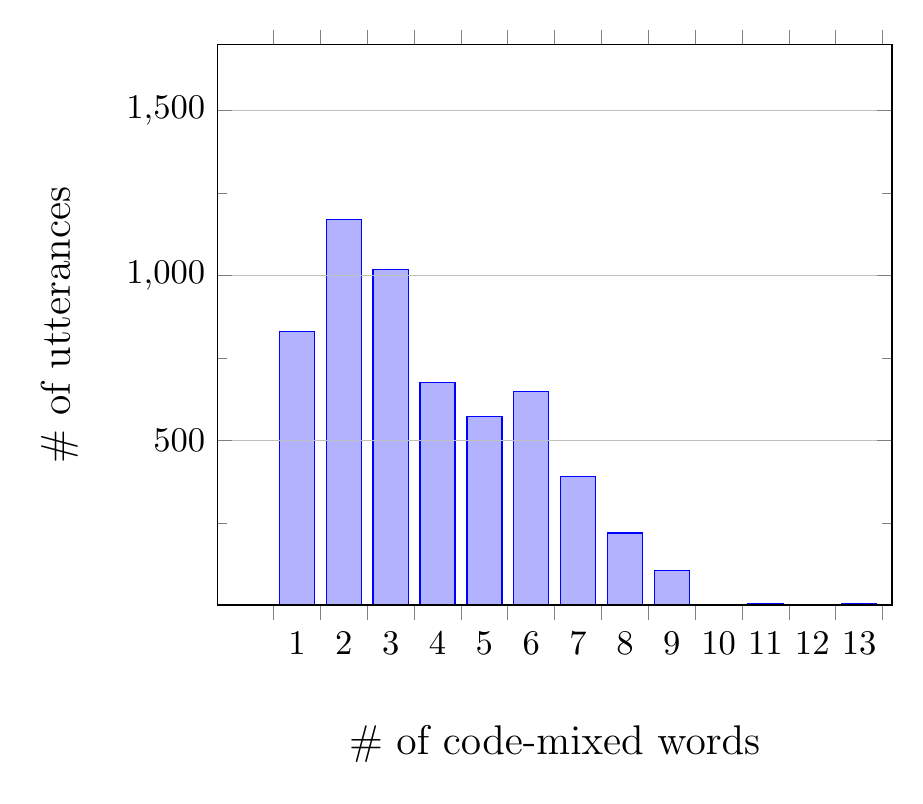}
            \caption[]%
            {{\small Gujarati}}    
            \label{guj_hist}
        \end{subfigure}
        \begin{subfigure}[b]{0.4\textwidth}   
            \centering 
            \includegraphics[width=0.92\textwidth]{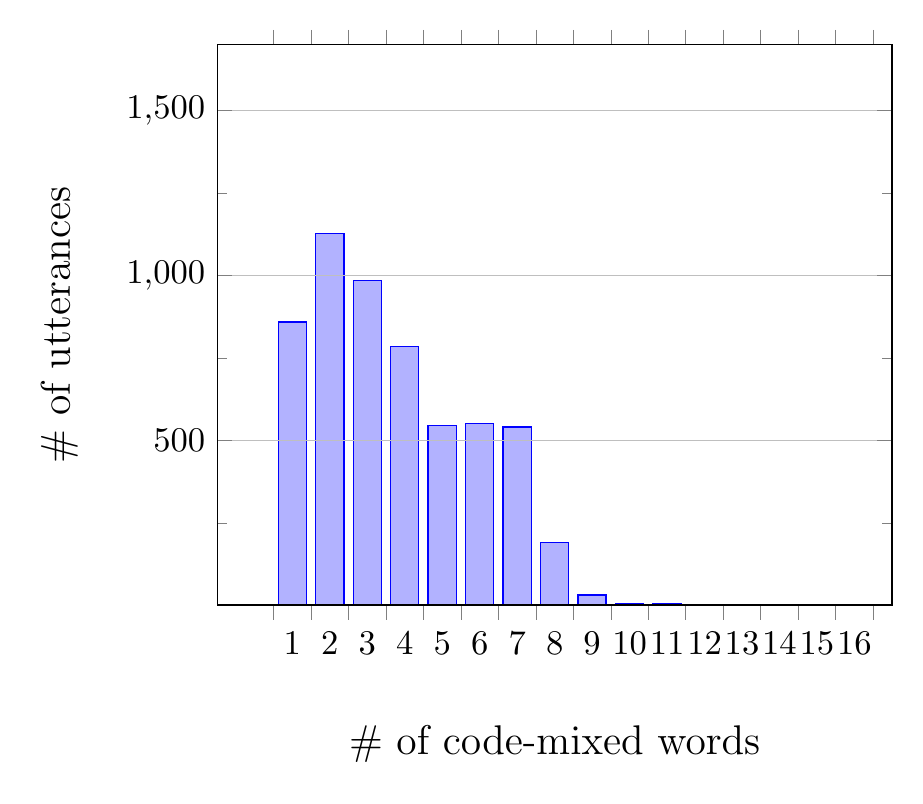}
            \caption[]%
            {{\small Tamil}}    
            \label{tam_hist}
        \end{subfigure}
        \caption[]
        {\label{cmhist} Histogram of the number of code mixed words in all the unique utterances for each language.} 
 \end{figure}

The percentage of code-mixed English words in the vocabulary of Hi-DSTC2, Be-DSTC2, Gu-DSTC2 and Ta-DSTC2 are 23.03\%, 26.24\%, 20.83\% and 19.40\% respectively. From these English words, the most high frequency words across all the four versions of the dataset were \textit{restaurant, food, town} and \textit{serve}. Although these words have their own dedicated counterparts in all the other four languages, people colloquially use these code-mixed English words very often when talking about restaurants in their native language. The percentage of  code-mixed utterances out of all the unique utterances in Hi-DSTC2, Be-DSTC2, Gu-DSTC2 and Ta-DSTC2 are 87.80\%, 90.90\%, 87.94\% and 84.49\% respectively (from Table \ref{cmstats}). This shows that a significant portion of the dataset contains code-mixed utterances and very few utterances are in pure native languages or in pure English. This fact is also evident from the average number of code-mixed utterances per dialog in Table \ref{cmstats} compared to the average number of utterances per dialog in Table \ref{enstats}. We also calculated the number of utterances which contain $k$ non-native words and then plotted a histogram (Figure \ref{cmhist}) where the x-axis shows the number of $k$ non-native words and the y-axis shows the number of utterances which had $k$ non-native words. 
These histograms show a similar trend across all the languages. Apart from such intra-utterance code-mixing, we also noticed some intra-word code-mixing in mostly Bengali (\textit{restauranter, towner}) and Tamil (\textit{addressum, numberum, addressah}) versions of the dataset. 

\begin{table}[]
\centering
\small{
\label{my-label}
\begin{tabular}{lllll}
\hline
                                                   & Hindi & Bengali & Gujarati & Tamil \\ \hline
Vocabulary Size                                    & 1676  &1372     & 1858     &2185   \\
Code-Mixed English Vocabulary                      &386   & 360    & 387     &424    \\
Native Language Vocabulary                         &739  &477      &912      &1214  \\
Others Vocabulary                                  &551   &535      & 559     &547  \\
Unique Utterances                                  &6549  & 6274    &6417      &6666  \\
Utterances with code-mixed words                   &5750   & 5703  & 5643   & 5632 \\
Pure Native Language utterances                    &348    &210     &340      &420 \\
Pure English utterances                            &451    & 361    & 434     &614 \\
Average length of utterances                       &8.16  &7.74   &8.04     &6.78  \\
Average \# of code-mixed utterances per dialog     &12.11  &14.28  &11.80  &12.96  \\
\hline
\end{tabular}
\caption{\label{cmstats} Statistics of the code-mixed dataset}}
\end{table}


\subsection{Quantitative Measures of Code-Mixing}
\newcite{CMI} introduced a measure to quantify the amount of code-mixing in a sentence as:

\begin{equation}
\centering
\label{base}
C_u(x) =\left\{
\begin{array}{ll}	
     \frac{N(x)-\underset{L_{i} \in \mathcal{L} }\max \{ t_{L_i} \}}{N(x)} &: N(x)>0\\
     0 &: N(x)=0
\end{array}\right. 
\end{equation}
\\
Here, $\mathcal{L}$ is the set of all languages in the corpus, $t_{L_i}$ is the number of tokens of language ${L_i}$ in the given sentence $x$, $\underset{L_{i} \in \mathcal{L} }\max \{ t_{L_i} \}$ is the maximum number of tokens of a language $L_i$ in the sentence $x$ and $N(x)$ is the number of language-specific tokens in the sentence ($N(x)$ does not include named entities as they are language agnostic). The authors make a crucial assumption that $\underset{L_{i} \in \mathcal{L} }\argmax \{ t_{L_i} \}$ is the \textit{Matrix} language and hence the numerator of Equation \ref{base} gives the number of foreign language tokens in $x$. This measure does not take into account the number of language switch points in a sentence (denoted by $P(x)$) and so the authors modify it further:

\begin{equation}
\centering
\label{cuP}
C_u(x) = 100\cdot\frac{N(x) - \underset{L_{i} \in \mathcal{L} }\max \{ t_{L_i} \} + P(x)}{2N(x)}: (\textit{if}~N(x) >0)
\end{equation}
The code-mixing in the entire corpus can then be quantified by taking an average of the above measure across all sentences in the corpus:
\begin{equation}
\centering
\label{c_avg}
C_{avg} = \frac{1}{U}\sum_{i=1}^{U}C_u(x_i)
\end{equation}
where $U$ is the number of sentences in the corpus. 
However, their main assumption that the language which has the maximum number of tokens in a sentence is the \textit{Matrix} language, may not always hold. Consider a counter example: ``Prezzo ek accha \textit{restaurant} hain \textit{in the north part of town} jo \textit{tasty chinese food serve} karta hain." Here the word `Prezzo' is a named entity and hence treated as a language independent token. The most frequent language (\textit{italicized}) is English but the \textit{Matrix} language is essentially Hindi. So we propose a small modification to their measure and replace $\underset{L_{i} \in \mathcal{L} }\max \{ t_{L_i} \}$ by the following:

\begin{equation}
\mathit{native}(x) = \left\{
\begin{array}{c l}	
     t_{L_n} &:  t_{L_n} > 0 \\
     N(x) &: t_{L_n} = 0
\end{array}\right. 
\end{equation}
where $L_n$ is the native (\textit{Matrix}) language of the utterance and $t_{L_n}$ is the number of tokens of the native language in the utterance. Note that we know the native (\textit{Matrix}) language of every utterance beforehand because of the manner in which the dataset was created. \newcite{CMI} also pointed out that $C_{avg}$ does not take the inter-utterance code-mixing and frequency of code-mixed utterances into account. To overcome this they proposed to use a term $\delta(x_i)$ which assigns a score of 1 if the \textit{Matrix} language of $x_i$ is different from that of $x_{i-1}$ or a score of 0 if they are same or $i=1$. Note that in our case $\delta(x_i)$ would mostly be 0 except for cases where $x_i$ is a pure English utterance. The authors also used a term for the fraction of code-mixed utterances $(\frac{S}{U})$ in the corpus, where $S$ is the total number of code-mixed utterances. We use a modified version of their final Code-Mixing index\footnote{We refer the reader to \newcite{CMI} for the detailed derivation.} by replacing the maximum function by $\mathit{native}(x)$:

\begin{equation}
C_c=\frac{100}{U}\left[ \frac{1}{2} \sum_{i=1}^{U} \left( 1-\frac{\mathit{native}(x)+P(x)}{N(x)} + \delta(x) \right) + \frac{5}{6}S\right]
\end{equation}

Similarly, \newcite{I-index} introduced the I-index measure to quantify the integration of different languages in a corpus. This metric is much simpler and simply computes the number of switch points in the corpus. For example, if a corpus contains $n$ words and there are $k$ positions at which the language of $word_i$ is not the same as the language of $word_j$ then the I-index is given by $\frac{k}{n-1}$. We compute the I-index for every utterance in a dialog, then compute the average over all utterances in a dialog and finally report the average across all dialogs in the code-mixed corpus. These measures of our dataset are shown in Table \ref{cmmix} and are compared with that of the existing datasets \cite{jamatia,vyas}. \newcite{jamatia} collected the code-mixed text from Twitter (TW) and Facebook (FB) posts whereas \newcite{vyas} collected their dataset only from Facebook forums. Although the dataset of \newcite{vyas} show the highest inter-utterance code-mixing ($\delta$), Hi-DSTC2 and Ta-DSTC2 show the highest level of overall code-mixing at the utterance level ($C_{avg}$) and the corpus level ($C_c$) respectively. 

\begin{table}[]
\centering
\small{
\begin{tabular}{|l|cccc|c|cccc|}
\hline
\textbf{\begin{tabular}[c]{@{}l@{}}Language-\\ pair\end{tabular}}   & \textbf{En-Be}                                         & \textbf{En-Hi}                                         & \textbf{En-Hi}                                         & \textbf{En-Hi}                                            & \textbf{En-Hi} & \textbf{En-Hi}                                      & \textbf{En-Be}                                      & \textbf{En-Gu}                                      & \textbf{En-Ta}                                      \\ \hline
\textbf{\begin{tabular}[c]{@{}l@{}}Dataset \end{tabular}} & \begin{tabular}[c]{@{}l@{}}TW\\ (Jamatia)\end{tabular} & \begin{tabular}[c]{@{}l@{}}TW\\ (Jamatia)\end{tabular} & \begin{tabular}[c]{@{}l@{}}FB\\ (Jamatia)\end{tabular} & \begin{tabular}[c]{@{}l@{}}FB+TW\\ (Jamatia)\end{tabular} & Vyas           & \begin{tabular}[c]{@{}l@{}}Hi-\\ DSTC2\end{tabular} & \begin{tabular}[c]{@{}l@{}}Be-\\ DSTC2\end{tabular} & \begin{tabular}[c]{@{}l@{}}Gu-\\ DSTC2\end{tabular} & \begin{tabular}[c]{@{}l@{}}Ta-\\ DSTC2\end{tabular} \\ \hline
\textbf{I-index}                                                    & -                                                      & -                                                      & -                                                      & -                                                         & -              & \textbf{0.04}                                                & \textbf{0.04}                                       & 0.03                                                & 0.03                                                \\ 
$\mathbf{C_{avg}}$                                                     & 8.34                                                   & 21.19                                                  & 3.92                                                   & 11.82                                                     & 11.44          & \textbf{32.12}                                      & 31.80                                               & 31.66                                               & 29.54                                               \\ 
$\mathbf{\delta}$                                                      & 22.09                                                  & 30.99                                                  & 6.70                                                   & 17.81                                                     & \textbf{53.50} & 26.38                                               & 29.06                                               & 24.50                                               & 38.32                                               \\ 
$\mathbf{C_c}$                                                       & 25.14                                                 & 64.38                                                  & 16.76                                                  & 38.53                                                     & 31.31          & 73.31                                               & 76.27                                               & 71.63                                      & \textbf{80.49}                                              \\ \hline
\end{tabular}
\caption{ \label{cmmix}
Comparison of the quantitative measures of code-mixing in the dataset.}}
\end{table}

\section{Baseline Models}
We establish some initial baseline results on this code-mixed dataset by evaluating two different generation based models: (i) sequence-to-sequence with attention \cite{seq2seq} and (ii) Hierarchical Recurrent Encoder-Decoder (HRED) model \cite{HRED}. Due to lack of space we don't describe these popular models here but refer the reader to the original papers. Apart from the above models, models which fetch the correct response from a set of candidate responses such as Query Reduction Networks \cite{QRN}, Memory Networks \cite{weston2017} and Hybrid Code Networks \cite{HCN} have also been evaluated on En-DSTC2. However, it is difficult to get candidate responses for every domain in practice and hence we stick to generation based models.
\subsection{Experimental Setup} We use the train, validation and test splits of \newcite{weston2017} mentioned in Table \ref{enstats}. We create training instances from the dialogs by creating pairs of \textit{\{context, response\}} where \textit{response} is every even numbered utterance and \textit{context} contains all the previous utterances. Thus, if a dialog has 10 utterances, we create 5 training instances from it. Similarly at the test time the model is given the \textit{context} and it has to generate the \textit{response}. For both the models, we used Adam optimizer \cite{adam} to train the network with a mini batch size of 32. We used dropouts \cite{dropout} of 0.25 and 0.35, initial learning rate of 0.0004 and Gated Recurrent Units (GRU) \cite{GRU} with hidden dimensions of size 350. We used word embeddings of size 300 with Glorot initialization \cite{xavier}. We also clipped the gradients at a maximum norm of 10 to avoid exploding gradients. 

\subsection{Evaluation}
We evaluate the performance of the above models using BLEU-4 \cite{bleu}, ROUGE-1, ROUGE-2 and ROUGE-L \cite{rouge} which are widely used to evaluate the performance of Natural Language Generation systems. We also compute the per utterance accuracy (exact match) by comparing the generated response with the ground truth response. The generated response is considered to be accurate only if it exactly matches the ground truth response. This is obviously a more strict metric for generation based models \cite{manning}. We also compute the per dialog accuracy by matching all the generated responses in a dialog with all the ground truth responses for that dialog. This metric measures whether the model was able to produce the entire dialog correctly end-to-end and hence complete the goal. We summarize the performance of the two models in Table \ref{basline}. We observe that the performance of these models is very similar across all the languages. We observe that the models are still far from 100\% accuracy and there is clearly scope for further improvement. 

\begin{table}[]
\centering
\small{
\begin{tabular}{lllllllllll}
\specialrule{.15em}{1em}{0em} 
            \multicolumn{1}{|c}{}& \multicolumn{5}{c}{\textsc{Seq2seq with Attention}}        & \multicolumn{5}{|c|}{\textsc{HRED}}           \\ \hline
           \multicolumn{1}{|c}{Metrics} &\multicolumn{1}{|c}{English} &Hindi & Bengali & Gujarati & Tamil &\multicolumn{1}{|c}{English}&Hindi & Bengali & Gujarati & \multicolumn{1}{c|}{Tamil} \\ \hline
\multicolumn{1}{|c}{BLEU-4}      &\multicolumn{1}{|c}{56.6} &  54.0    &56.8         &53.8          & 62.1      &\multicolumn{1}{|c}{57.8}       &54.1         &56.7  & 54.1         &\multicolumn{1}{c|}{60.7}      \\
\multicolumn{1}{|c}{ROUGE-1}     &\multicolumn{1}{|c}{67.2} &62.9      &67.4        & 64.7         & 67.8      & \multicolumn{1}{|c}{67.9}      &63.3         &67.1& 65.3         & \multicolumn{1}{c|}{67.1}      \\
\multicolumn{1}{|c}{ROUGE-2}     &\multicolumn{1}{|c}{55.9}    &52.4   &57.5         & 54.8         &56.3       & \multicolumn{1}{|c}{57.5}      &52.6         &56.9&55.2          & \multicolumn{1}{c|}{55.6}      \\
\multicolumn{1}{|c}{ROUGE-L}     &\multicolumn{1}{|c}{64.8}    & 61.0  &65.1         &62.6          &65.6       & \multicolumn{1}{|c}{65.7}      &61.5         &64.8&63.2          &\multicolumn{1}{c|}{65.1}       \\
\multicolumn{1}{|c}{Per response acc.} &\multicolumn{1}{|c}{46.0} &48.0      &50.4         &47.6          &49.3       & \multicolumn{1}{|c}{48.8}      &47.2         &47.7&47.9          & \multicolumn{1}{c|}{47.8}   \\
\multicolumn{1}{|c}{Per dialog acc.} &\multicolumn{1}{|c}{1.4} &1.2      &1.5         &1.5          &1.3       & \multicolumn{1}{|c}{1.4}      &1.5         & 1.6&1.6          & \multicolumn{1}{c|}{1.0}   \\\hline  
\end{tabular}
\caption{\label{basline}Performance of the baseline models on all the languages}}
\end{table}

\section{Conclusion} 
Code-mixing is an emerging trend of communication in the multilingual regions. The community has already addressed this phenomenon by introducing challenges on POS-Tagging, Language Identification, Language Modeling, \textit{etc} on the code-mixed corpora. However, the approaches to development of dialog systems still rely on monolingual conversation datasets. To alleviate this problem we introduced a goal-oriented code-mixed dialog dataset for four languages (Hindi-English, Bengali-English, Gujarati-English and Tamil-English respectively). The dataset was created using a mix of in-house and crowdsourced workers. All the utterances in the dataset were evaluated by in-house evaluators and the overall dialogs were also evaluated for colloquialism, intelligibility and coherence. On all these measures, the dialogs in our dataset received a high score. To facilitate further research on these datasets, we provide the implementation of two popular neural dialog models \textit{viz.} sequence-to-sequence  and HRED. The evaluation of these models suggest that there is a clear scope for development of new architectures which can understand and converse in code-mixed languages.


\section*{Acknowledgements}
We would like to thank  Accenture Technology Labs, India for supporting this work through their generous academic research grant.


 \newpage
\section*{Appendix A. Instructions to Crowdsourced Workers}
\begin{figure}[!ht]
\resizebox{\linewidth}{!}{
\centering
\includegraphics[scale=0.011]{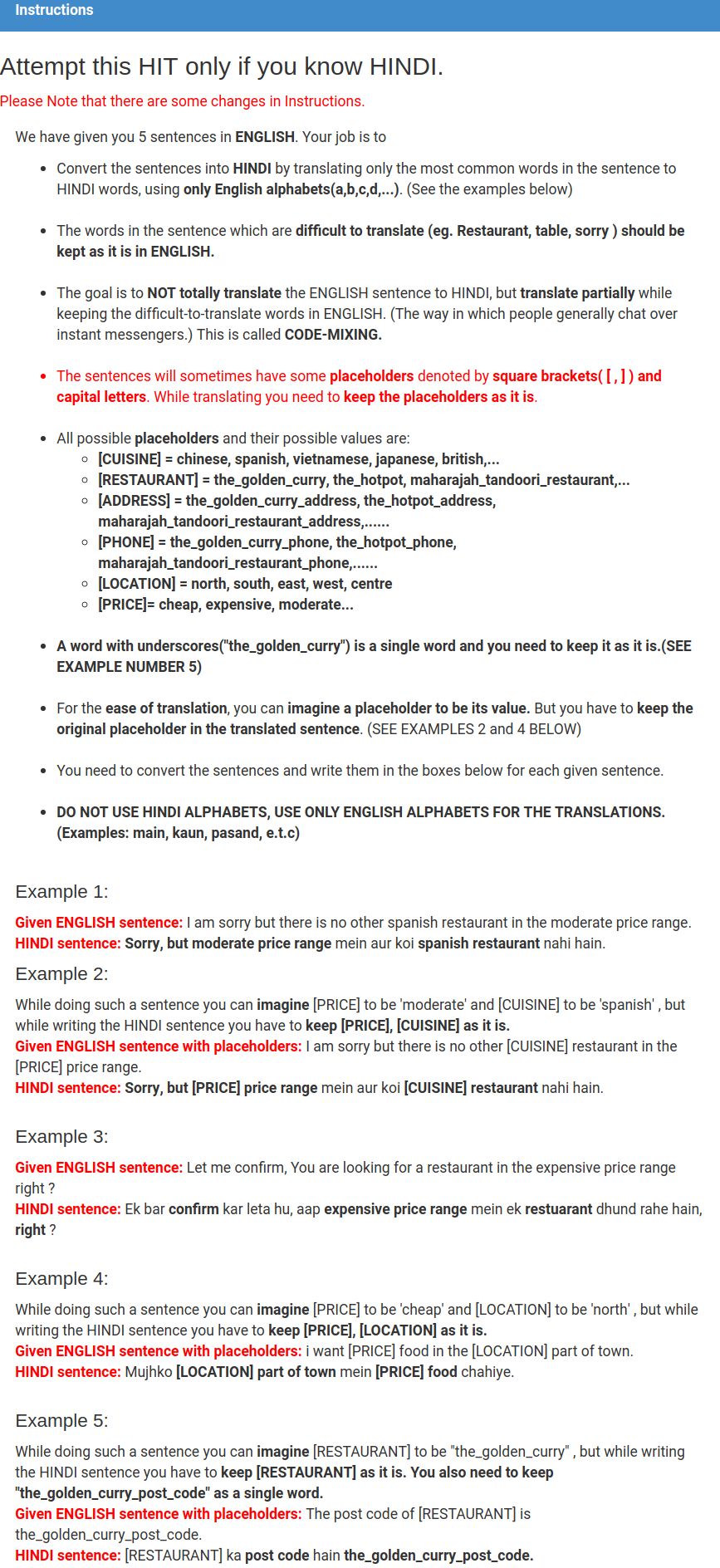}
}
\caption{Screenshot of the instructions and examples provided to the crowdsourced workers}
\end{figure}

\section*{Appendix B. Sample Code-Mixed Chats from the Dataset}

\begin{table}[!ht]
\centering
\small{
\begin{tabular}{cl}
\hline
\multicolumn{1}{c|}{User}                                          &\multicolumn{1}{c}{Bot}                                                                                                                                                                            \\ \hline
 \multicolumn{1}{l|}{1. \textless SILENCE\textgreater}                  & \begin{tabular}[c]{@{}l@{}}Hello , Cambridge restaurant system mein aapka \\ swagat hai.  Aap pa sakte hai restaurant by area , \\price range ya food type. Aap ko kaise help kar \\sakta hu main ?\end{tabular} \\\hline
 \multicolumn{1}{l|}{\begin{tabular}[c]{@{}l@{}}2. main north part of town mein\\ \quad cheap restaurant dhund raha hoon \end{tabular}  } & Aap kaunse type ka food lena chahege?                                                                                                                                                  \\\hline
 \multicolumn{1}{l|}{3. mujhe parvah nahi hain.}                              & api\_call R\_cuisine north cheap                                                                                                                                            \\\hline

\multicolumn{2}{l}{\qquad da\_vinci\_pizzeria R\_post\_code da\_vinci\_pizzeria\_post\_code}                                                                                                                                                       \\
\multicolumn{2}{l}{\qquad da\_vinci\_pizzeria R\_cuisine italian}                                                                                                                                                                                         \\
\multicolumn{2}{l}{\qquad da\_vinci\_pizzeria R\_location north}                                                                                                                                                                                          \\
\multicolumn{2}{l}{\qquad da\_vinci\_pizzeria R\_phone da\_vinci\_pizzeria\_phone}                                                                                                                                                                 \\
\multicolumn{2}{l}{\qquad da\_vinci\_pizzeria R\_address da\_vinci\_pizzeria\_address}                                                                                                                                                             \\
\multicolumn{2}{l}{\qquad da\_vinci\_pizzeria R\_price cheap}                                                                                                                                                                                          \\
\multicolumn{2}{l}{\qquad da\_vinci\_pizzeria R\_rating 10}                                                                                                                                                                                                \\
\multicolumn{2}{l}{\qquad royal\_spice R\_post\_code royal\_spice\_post\_code}                                                                                                                                                                 \\
\multicolumn{2}{l}{\qquad royal\_spice R\_cuisine indian}                                                                                                                                                                                             \\
\multicolumn{2}{l}{\qquad royal\_spice R\_location north}                                                                                                                                                                                               \\
\multicolumn{2}{l}{\qquad royal\_spice R\_phone royal\_spice\_phone}                                                                                                                                                                           \\
\multicolumn{2}{l}{\qquad royal\_spice R\_address royal\_spice\_address}                                                                                                                                                                       \\
\multicolumn{2}{l}{\qquad royal\_spice R\_price cheap}                                                                                                                                                                                               \\
\multicolumn{2}{l}{\qquad royal\_spice R\_rating 8}                                                                                                                                                                                                    \\\hline
 \multicolumn{1}{l|}{5. \textless SILENCE\textgreater}                  & \begin{tabular}[c]{@{}l@{}}north of town me da\_vinci\_pizzeria \\ek achaa jagah hai aur prices cheap hai\end{tabular}                                              \\\hline
 \multicolumn{1}{l|}{6. aur kuch}                        & \begin{tabular}[c]{@{}l@{}}north of town me royal\_spice \\ek achaa jagah hai aur prices cheap hai\end{tabular}                                                                          \\\hline
 \multicolumn{1}{l|}{7. kyaa mujhe phone number mil sakta hai}           & \begin{tabular}[c]{@{}l@{}}royal\_spice ka phone number \\hai royal\_spice\_phone\end{tabular}                                                                  \\\hline
 \multicolumn{1}{l|}{8. thank you goodbye}                       & Welcome , alvida                                                                    \\\hline                                                                                                 
\end{tabular}
}
\caption{\label{example_chat} An example chat from the Hindi version of DSTC2 dataset}
\end{table}

\begin{table}[!ht]
\centering
\small{
\begin{tabular}{cl}
\hline
\multicolumn{1}{c|}{User}                                          &\multicolumn{1}{c}{Bot}                                                                                                                                                                            \\ \hline
 \multicolumn{1}{l|}{1. \textless SILENCE\textgreater}                  & \begin{tabular}[c]{@{}l@{}}Hello, Cambridge restaurant system ey apnake \\swagoto janai. Apni restaurants khujte paren food \\type, price ar area diye. Apnake ki bhabe help\\ korte pari ami ?\end{tabular} \\\hline
 \multicolumn{1}{l|}{\begin{tabular}[c]{@{}l@{}}2. Ami ekta cheap restaurant khujchi \\ \quad town er north part e \end{tabular}  } & Kon type er khabar apni pochondo korben ?                                                                                                                                                 \\\hline
 \multicolumn{1}{l|}{3. ami care kori na}                              & api\_call R\_cuisine north cheap                                                                                                                                            \\\hline

\multicolumn{2}{l}{\qquad da\_vinci\_pizzeria R\_post\_code da\_vinci\_pizzeria\_post\_code}                                                                                                                                                       \\
\multicolumn{2}{l}{\qquad da\_vinci\_pizzeria R\_cuisine italian}                                                                                                                                                                                         \\
\multicolumn{2}{l}{\qquad da\_vinci\_pizzeria R\_location north}                                                                                                                                                                                          \\
\multicolumn{2}{l}{\qquad da\_vinci\_pizzeria R\_phone da\_vinci\_pizzeria\_phone}                                                                                                                                                                 \\
\multicolumn{2}{l}{\qquad da\_vinci\_pizzeria R\_address da\_vinci\_pizzeria\_address}                                                                                                                                                             \\
\multicolumn{2}{l}{\qquad da\_vinci\_pizzeria R\_price cheap}                                                                                                                                                                                          \\
\multicolumn{2}{l}{\qquad da\_vinci\_pizzeria R\_rating 10}                                                                                                                                                                                                \\
\multicolumn{2}{l}{\qquad royal\_spice R\_post\_code royal\_spice\_post\_code}                                                                                                                                                                 \\
\multicolumn{2}{l}{\qquad royal\_spice R\_cuisine indian}                                                                                                                                                                                             \\
\multicolumn{2}{l}{\qquad royal\_spice R\_location north}                                                                                                                                                                                               \\
\multicolumn{2}{l}{\qquad royal\_spice R\_phone royal\_spice\_phone}                                                                                                                                                                           \\
\multicolumn{2}{l}{\qquad royal\_spice R\_address royal\_spice\_address}                                                                                                                                                                       \\
\multicolumn{2}{l}{\qquad royal\_spice R\_price cheap}                                                                                                                                                                                               \\
\multicolumn{2}{l}{\qquad royal\_spice R\_rating 8}                                                                                                                                                                                                    \\\hline
 \multicolumn{1}{l|}{5. \textless SILENCE\textgreater}                  & \begin{tabular}[c]{@{}l@{}}da\_vinci\_pizzeria ekta nice jayega \\town er north ey jekhane prices cheap\end{tabular}                                              \\\hline
 \multicolumn{1}{l|}{6. ar onno kichu ache}                        & \begin{tabular}[c]{@{}l@{}}royal\_spice ekta nice jayega town \\er north ey jekhane prices cheap\end{tabular}                                                                          \\\hline
 \multicolumn{1}{l|}{7. ami ki phone number ta pete pari}           & \begin{tabular}[c]{@{}l@{}}royal\_spice er phone number holo\\ royal\_spice\_phone\end{tabular}                                                                  \\\hline
 \multicolumn{1}{l|}{8. thank you goodbye}                       & you are welcome                                                                  \\\hline                                                                                                 
\end{tabular}
}
\caption{\label{example_chat} An example chat from the Bengali version of DSTC2 dataset}
\end{table}

\end{document}